\documentclass[runningheads]{llncs}
\usepackage[T1]{fontenc}
\usepackage{graphicx}
\usepackage{booktabs}
\usepackage[misc]{ifsym}
\usepackage{booktabs} 
\usepackage{amsmath,amssymb}
\usepackage{longtable}
\usepackage{graphicx}
\usepackage{subcaption}
\usepackage{xcolor}

\usepackage{tocloft}
\usepackage{titletoc}
\usepackage{array, tabularx, booktabs, multicol}
 \usepackage{multirow}

\usepackage[numbers]{natbib}

\newcommand{\corr}{(\Letter)}
\usepackage{mwe}

\begin{document}

\title{STN-TGAT: Top-K Portfolio Construction via Prior-Guided Graph Attention with Learnable Soft-Threshold Sparsification}

\titlerunning{STN-TGAT}

\author{Haoran Guo\inst{1} \and
Yutong Lu\inst{2} \and
Li Zhang \corr \inst{1}}
\authorrunning{H. Guo et al.}
\institute{
University College London, Institute of Financial Technology, London, UK\\
\and
University of Oxford, Oxford, UK,
\email{\{haoran.guo.24,ucesl07\}@ucl.ac.uk},\email{yutong.lu@institute2.ac.uk}
}
\footnotetext{Corresponding author: Li Zhang (\texttt{ucesl07@ucl.ac.uk})}

\maketitle              

\begin{abstract}

This paper tackles the problem of stock ranking and portfolio construction under realistic investment settings by jointly modeling temporal dynamics and cross-sectional dependencies. We propose the Soft-Threshold NMI-prior Transformer Graph Attention Network (STN-TGAT), which integrates a temporal Transformer with a Graph Attention Network to capture long-horizon sequential patterns and dynamic inter-stock relationships. An NMI-based prior graph combined with a soft-threshold sparsification mechanism enhances structural robustness by mitigating noisy correlations while preserving informative connections. The portfolio formation process incorporates practical considerations, including Top-5 selection within the Top-50 $S\&P$ 500 constituents, explicit weight allocation, and transaction cost adjustment, thereby aligning the evaluation with real-world trading conditions. Empirical results on real-world data demonstrate that STN-TGAT consistently outperforms benchmark models from predictive accuracy and investment profitability measured by portfolio returns. These findings suggest that combining decision-aligned training with adaptive relational modeling provides a coherent and practically effective framework for data-driven portfolio construction.

\end{abstract}

\section{Introduction}
Equity portfolio construction in quantitative asset management is naturally formulated as a repeated Top-$K$ selection problem. On each trading day, an investable universe is ranked and capital is allocated to a small subset of the highest-ranked assets under practical constraints such as transaction costs and position limits. In this decision-centric setting, minimizing prediction error alone is insufficient; the primary objective is achieving reliable ranking performance at the top of the list, as these assets determine realized portfolio returns.

Daily financial markets present substantial challenges, including noise, nonlinear dynamics, volatility clustering, and regime shifts. Traditional econometric models such as ARIMA and GARCH capture autoregressive structure and conditional heteroskedasticity \cite{patel2024systematic}, but rely on linear assumptions that limit their ability to represent complex market dependencies \cite{yu2024stock}. Deep learning architectures, including LSTM and GRU, improve temporal representation by modeling longer horizons \citep{gao2021graph}. However, many temporal approaches treat each stock independently, overlooking cross-sectional dependencies induced by sector exposure, style factors, and shared macroeconomic shocks \citep{jafari2022gcnet}.

Graph Neural Networks (GNNs) provide a natural mechanism to incorporate such relationships by representing stocks as nodes and dependencies as edges. Hybrid architectures combining Transformers with Graph Attention Networks (GATs) aim to jointly model temporal dynamics and relational structure. Yet, their effectiveness depends critically on graph construction. Dense graphs may propagate noise, while overly sparse or heuristically thresholded graphs risk discarding informative dependencies. These challenges motivate adaptive mechanisms for learning structured cross-sectional relationships.


To address these challenges, we propose Soft-Thresholded NMI-prior Transformer Graph Attention Network  (STN-TGAT), a decision-oriented framework that integrates self-attention-based temporal modeling with adaptive graph learning. Instead of predicting absolute returns, the model directly learns to rank stocks by relative performance, aligning the learning objective with daily portfolio construction where only the top candidates are selected. The framework includes two key components. First, we introduce a top-weighted ListNet ranking objective that emphasizes ordering accuracy among the top predictions, reflecting the practical importance of top-ranked assets in a Top-$K$ selection strategy. Second, we incorporate an information-theoretic relational prior based on Normalized Mutual Information (NMI) and apply a learnable soft-threshold sparsification mechanism to adaptively control graph density. This design aims to suppress noisy relations while preserving informative co-movement structures.

This paper makes three contributions:
\begin{enumerate}
\item We formulate daily equity selection as a Top-$K$ ranking problem aligned with portfolio construction and introduce a top-weighted ListNet objective that emphasizes accuracy among the highest-ranked assets.
\item We develop a prior-guided adaptive graph that integrates an NMI-based relational prior with learnable soft-threshold sparsification to refine cross-sectional dependency modeling.
\item We conduct evaluation using both ranking metrics and realistic net-of-fee portfolio backtesting with transaction costs, linking predictive performance to economically meaningful outcomes.
\end{enumerate}


\section{Related Work}
\subsection{Time-series modeling for financial prediction}

Early financial forecasting studies relied on statistical models such as ARIMA and GARCH to capture autoregressive dynamics and volatility patterns in financial time series \citep{pande2025forecasting}. Although effective for short-term dynamics, these models rely largely on linear assumptions and struggle to represent the nonlinear and regime-dependent behavior of modern markets. Classical machine learning approaches, including Support Vector Machines (SVMs) and Random Forests (RFs), were later introduced to incorporate richer feature representations \citep{ren2018forecasting, illa2022stock, omar2022stock}. However, these methods often ignore long-range temporal dependencies and remain sensitive to non-stationarity \cite{patel2024systematic}.

Deep learning models have substantially improved temporal representation learning. Recurrent architectures such as LSTM and GRU capture multi-horizon dependencies in financial sequences \citep{feng2022relation, pande2025forecasting}, while hybrid architectures combine convolutional layers, recurrent networks, and attention mechanisms to extract both local patterns and long-term dynamics \citep{chen2022dynamic, cui2023novel, pham2024improved}. More recently, Transformer-based models have gained prominence for their ability to model long-range dependencies through self-attention \citep{wang2022stock}, with extensions including hierarchical or frequency-aware variants \citep{ding2020hierarchical, li2022stock}. Nevertheless, most existing approaches emphasize pointwise prediction accuracy rather than the relative ranking and Top-$K$ selection required in practical portfolio construction.

\subsection{Graph neural networks for cross-sectional dependency learning}

Graph neural networks (GNNs) provide a natural framework for modeling cross-sectional dependencies among stocks \cite{you2024dgdnn}. By representing equities as nodes and their relationships as edges, GNNs enable information propagation across related assets. Early studies used static graphs combined with graph convolution to capture stable relationships such as sector membership or persistent co-movement \citep{jafari2022gcnet}. However, static graph structures often fail to reflect the evolving dependencies and regime shifts common in financial markets.

To improve flexibility, attention-based GNNs dynamically learn neighbor importance and can be integrated with temporal encoders. For example, RA-GAT combines LSTM-based temporal representations with graph attention to model inter-stock relationships \citep{feng2022relation}, while other spatio-temporal architectures jointly learn temporal and relational signals for stock prediction or ranking tasks \citep{zheng2023relational, yin2021forecasting}. Extensions such as multi-graph models and dynamic graph learning further capture heterogeneous relations or evolving network structures \citep{wang2022mg, ma2024vgc, tian2023learning,zhang2020feature,zhang2020hophop_arxiv}. These studies demonstrate the value of relational modeling, while also highlighting the sensitivity of GNN performance to the quality and sparsity of the underlying graph.

\subsection{Dependence measures and graph construction}

A key step in graph-based financial modeling is defining meaningful inter-stock dependencies \cite{you2024dgdnn}. Linear correlation measures such as Pearson correlation are widely used due to their simplicity \citep{castilho2024forecasting, gu2024dystage}, while rank-based statistics like Spearman or Kendall correlations improve robustness under non-Gaussian distributions \citep{hu2025study}. To capture nonlinear relationships, alternative measures including distance correlation, normalized mutual information (NMI), and detrended cross-correlation analysis have been explored \citep{ugwu2023distance, turner2023graph, feng2022relation}. However, these measures typically produce dense dependency matrices that require additional sparsification.

Various filtering techniques, such as dynamic thresholding or minimal spanning tree structures, have been used to extract informative graph topology \citep{zhang2022node,castilho2024forecasting}. Nevertheless, heuristic sparsification strategies may be unstable under regime shifts and may not align well with downstream learning objectives. Motivated by these limitations, our work combines an NMI-based dependence prior with a learnable soft-threshold sparsification mechanism and integrates it with attention-based temporal encoding, enabling adaptive cross-sectional dependency modeling aligned with Top-$K$ portfolio selection.

\section{Task Formulation}

We formulate daily equity selection as a cross-sectional Top-$K$ ranking problem. 
For each trading day, we observe standardized input features for $N$ stocks over a lookback window of length $L$, denoted by $\mathbf{X} \in \mathbb{R}^{N \times L \times F}$, where $F$ is the feature dimension. The $i$-th stock’s temporal features are given by $\mathbf{X}_i \in \mathbb{R}^{L \times F}$. \footnote{
Throughout the paper, we omit explicit time indices (e.g., writing $\mathbf{X}$ instead of $\mathbf{X}_t$) for notational simplicity. 
Unless otherwise specified, all variables are implicitly indexed by trading day $t$, and the model operates sequentially 
in a rolling-window manner to predict $t{+}1$ returns.
}

The model maps $\mathbf{X}$ to stock-level scores $\mathbf{s} = (s_1, \dots, s_N) \in \mathbb{R}^N$, where each $s_i$ is produced by the spatio-temporal encoder and prediction head. These scores induce a ranking over the cross-section. Let $y_i$ denote the realized next-day log return of stock $i$ and let $\mathbf{y}^{\mathrm{rank}}$ denote the cross-sectionally standardized returns used as ranking targets. \footnote{The next-day log return is defined as $r_{i,t} = \log P_{i,t} - \log P_{i,t-1}$,
where $P_{i,t}$ denotes the adjusted closing price of stock $i$ on day $t$. $y_{i,t} = r_{i,t}$, $y^{\mathrm{rank}}_{i,t} =  \frac{r_{i,t} - \mu_{t}}{\sigma_{t}}$, where $\mu_{t}$ and $\sigma_{t}$ denote the cross-sectional mean and 
standard deviation of $\{ r_{j,t} \}_{j=1}^{N}$ across all stocks on day $t$.} Portfolio construction selects the Top-$K$ stocks according to predicted scores,
\[
\mathcal{S} = \operatorname{TopK}(\mathbf{s}, K),
\]
and allocates capital within $\mathcal{S}$ under transaction cost and rebalancing constraints. Therefore, the learning objective emphasizes ranking consistency particularly in the upper portion of the cross-section—while optionally preserving return magnitude information for allocation.

\section{Methodology}

\subsection{Framework Overview}

The proposed Soft-Thresholded NMI-prior Transformer Graph Attention Network (STN-TGAT) jointly models temporal dynamics and cross-sectional dependencies for Top-$K$ equity selection. A Transformer-based encoder extracts long-range sequential patterns from historical features and produces stock embeddings via attention pooling. These embeddings are refined through a Graph Attention Network guided by an NMI-based prior with learnable soft-threshold sparsification to adaptively control relational structure. Finally, a shared prediction head produces stock-level scores that are optimized using a head-weighted listwise ranking loss, optionally combined with a regression term to preserve magnitude information. During inference, stocks are ranked by predicted scores, and the top-$K$ assets are selected to form the portfolio under transaction cost adjustments. This design jointly integrates temporal modeling, adaptive relational learning, and decision-aligned optimization for practical Top-$K$ portfolio construction. For clarity, key notations used throughout the methodology are summarized in Appendix.

\subsection{Temporal Encoding}

Given the input tensor $\mathbf{X} \in \mathbb{R}^{N \times L \times F}$ defined in the task formulation, we encode each stock’s lookback sequence using an encoder-only Transformer \citep{vaswani2017attention}. A linear projection is first applied along the feature dimension independently for each stock and time step:
\begin{equation}
\bar{\mathbf{X}} = \mathbf{X}\mathbf{W}^{(I)} + \mathbf{PE},
\end{equation}
where $\mathbf{W}^{(I)}\in\mathbb{R}^{F\times d}$ is a learnable projection matrix. The sinusoidal positional encodings $\mathbf{PE}\in\mathbb{R}^{L\times d}$ are shared across stocks and added along the temporal dimension. The resulting representations are processed by a standard Transformer encoder, producing hidden states $\mathbf{H}\in\mathbb{R}^{N\times L\times d}$.

To obtain a fixed-dimensional representation for each stock, we apply attention-based temporal pooling. Let $\mathbf{h}_t\in\mathbb{R}^{d}$ denote the hidden state at time step $t$ ($t = 1, \dots, L$ indexes the lookback window).  With a learnable query vector $\mathbf{q}\in\mathbb{R}^{d}$, the attention weights are computed as
\begin{equation}
a_t = 
\frac{
\exp\left( \langle \mathbf{h}_t, \mathbf{q} \rangle \right)
}{
\sum_{k=1}^{L}
\exp\left( \langle \mathbf{h}_k, \mathbf{q} \rangle \right)
},
\qquad
\mathbf{z} = \sum_{t=1}^{L} a_t \mathbf{h}_t \in \mathbb{R}^{d}.
\end{equation}
Applying this pooling independently to each stock yields the cross-sectional daily representation $\mathbf{Z}\in\mathbb{R}^{N\times d}$, which serves as node features for the downstream graph module.

\subsection{Spatial Encoding}

\subsubsection{NMI-based Dependence Graph with Learnable Soft-Thresholding}

Following \citet{feng2022relation}, we construct a cross-sectional dependence graph from daily log-returns using Normalized Mutual Information (NMI), which captures nonlinear co-movement beyond linear correlation. Let $r_i=\{\Delta\log P_{t,i}\}_{t=1}^{T}$ denote the training-window return series of stock $i$ and $ P_{t,i}$ denotes the adjusted closing price of stock $i$ on day $t$. Mutual information is estimated using histogram-based equal-width binning with $k_b$ bins to obtain empirical marginal and joint distributions. The normalized mutual information between stocks $i$ and $j$ is computed as
\begin{equation}
\mathrm{NMI}(r_i,r_j)
=
\frac{2\,I(r_i;r_j)}{H(r_i)+H(r_j)}
\in [0,1],
\end{equation}
where $I(\cdot;\cdot)$ and $H(\cdot)$ denote the histogram-based estimates of mutual information and Shannon entropy, respectively. This yields a dense adjacency matrix $\mathbf{A}_0 \in \mathbb{R}^{N\times N}$ 
\begin{equation}
(\mathbf{A}_0)_{ij} = \mathrm{NMI}(r_i,r_j),
.
\end{equation}

Dense connectivity may increase the risk of over-smoothing during repeated neighborhood aggregation. To address this, we introduce a differentiable gating mechanism that adaptively attenuates weak edges. Specifically, we define a gate matrix $\mathbf{G}=[g_{ij}] \in (0,1)^{N\times N}$ as
\begin{equation}
g_{ij} = \sigma\!\left(\beta\big((\mathbf{A}_0)_{ij}-t_g\big)\right),
\end{equation}
where $\sigma(\cdot)$ denotes the sigmoid function, $\beta>0$ controls the sharpness of the transition, larger values make the gating function approximate a hard threshold, while smaller values yield smoother attenuation. $t_g\in(0,1)$ is a learnable threshold parameterized. The sparsified adjacency is obtained via element-wise modulation $\mathbf{A}_\phi = \mathbf{A}_0 \odot \mathbf{G}$. Edges with weights below $t_g$ are smoothly suppressed, while stronger connections are largely preserved. The threshold parameter is learned jointly with the overall training objective.

To preserve symmetry and stabilize scale, we symmetrize the learned adjacency as $\mathbf{A}_\phi \leftarrow \tfrac{1}{2}\big(\mathbf{A}_\phi + \mathbf{A}_\phi^\top\big),$
followed by global max-scaling, $\mathbf{A}_\phi \leftarrow \mathbf{A}_\phi / \max_{i,j} (\mathbf{A}_\phi)_{ij},$
ensuring $\mathbf{A}_\phi \in [0,1]^{N\times N}$. A small diagonal shift $\mathbf{A}_\phi \leftarrow \mathbf{A}_\phi + \delta \mathbf{I}$ is added to guarantee self-connections and avoid empty-neighborhood cases. The resulting adjacency $\mathbf{A}_\phi$ is used in the subsequent Graph Attention Network to guide spatial aggregation according to the learned relational structure.

\subsubsection{GAT with NMI Prior}

Given the cross-sectional embeddings 
$\mathbf{Z}\in\mathbb{R}^{N\times d}$ from the temporal encoder,
we apply a multi-head Graph Attention Network (GAT) \cite{velivckovic2017graph} to model inter-stock dependencies.
The learned sparsified adjacency $\mathbf{A}_\phi$ defines the graph structure.
For computational efficiency, each node $i$ aggregates information from its
$K_{\text{nbr}}$ strongest neighbors under $\mathbf{A}_\phi$, denoted by $\mathcal{N}(i)$.

Let $\mathbf{H}^{(0)}=\mathbf{Z}$.
For GAT layer $\ell$ and attention head $h$, we compute head-specific projections
\begin{equation}
\mathbf{H}^{(\ell,h)}=\mathbf{H}^{(\ell)} \mathbf{W}^{(\ell,h)},
\end{equation}
where $\mathbf{W}^{(\ell,h)}\in\mathbb{R}^{d\times d_h}$ are learnable parameters applied independently to each node representation.

Unlike standard GAT, we inject the NMI-based structural prior into the attention logits.
For $j\in\mathcal{N}(i)$,
\begin{equation}
e^{(\ell,h)}_{ij}
=
\frac{\mathrm{LeakyReLU}\!\left(
\mathbf{a}^{(\ell,h)\top}
[\mathbf{H}^{(\ell,h)}_i \,\|\, \mathbf{H}^{(\ell,h)}_j]
\right)}
{\tau_{\text{a}}}
+
\lambda \log\!\left((\mathbf{A}_\phi)_{ij}+\varepsilon\right),
\end{equation}
where $\mathbf{a}^{(\ell,h)}\in\mathbb{R}^{2d_h}$ is learnable,
$\tau_{\text{a}}>0$ controls attention sharpness,
and $\varepsilon>0$ ensures numerical stability.
The first term captures data-driven feature compatibility,
while the second term biases attention toward edges supported by the structural prior.
The prior strength is controlled by
$\lambda \in (0,1)$.

Attention weights are obtained via softmax over $\mathcal{N}(i)$:
\begin{equation}
\alpha^{(\ell,h)}_{ij}
=
\frac{\exp(e^{(\ell,h)}_{ij})}
{\sum_{k\in\mathcal{N}(i)} \exp(e^{(\ell,h)}_{ik})}.
\end{equation}

Node representations are updated using standard multi-head aggregation \citep{velivckovic2017graph}, followed by residual connections and layer normalization.
After $L_g$ layers, we obtain $\mathbf{H}^{(L_g)}\in\mathbb{R}^{N\times d}$. A linear prediction head is applied to each node representation:
\begin{equation}
s_i = \mathbf{w}_{\text{out}}^\top \mathbf{H}^{(L_g)}_i + b,
\qquad
\mathbf{s}\in\mathbb{R}^{N},
\label{eq:score_head}
\end{equation}
where $\mathbf{w}_{\text{out}}\in\mathbb{R}^{d}$ and $b\in\mathbb{R}$ are learnable parameters shared across stocks.


\subsection{Loss Function}

To align model training with downstream Top-$K$ portfolio construction, we optimize a composite objective consisting of three components: (i) a head-weighted listwise ranking loss, (ii) an auxiliary regression loss, and (iii) graph sparsity regularization. Direct optimization of portfolio utility is non-differentiable and unstable due to discrete Top-$K$ selection and transaction costs; therefore, we adopt differentiable ranking-based surrogates that approximate the decision structure while preserving tractability.

\paragraph{Head-Weighted ListNet Objective.}

Let $\mathbf{s}\in\mathbb{R}^N$ denote predicted stock scores, and let $\mathbf{y}^{\mathrm{rank}}\in\mathbb{R}^N$ denote cross-sectionally standardized next-period returns used as ranking targets.

Following ListNet, we define the student and teacher distributions:
\begin{equation}
P_i = \frac{\exp(s_i)}{\sum_{j=1}^{N}\exp(s_j)},
\qquad
Q_i = \frac{\exp(y^{\mathrm{rank}}_i / \tau_{\mathrm{rank}})}
{\sum_{j=1}^{N}\exp(y^{\mathrm{rank}}_j / \tau_{\mathrm{rank}})},
\end{equation}
where $\tau_{\mathrm{rank}}>0$ controls teacher sharpness. Cross-sectional standardization stabilizes numerical scaling.

Standard ListNet minimizes $\mathrm{KL}(Q\|P)$, aligning full ranking distributions. However, portfolio construction depends primarily on the highest-ranked assets. Let $\mathrm{pos}(i)$ denote the rank position of stock $i$ under the teacher distribution $Q$ (sorted in descending order). To emphasize accuracy in higher-ranked positions while preserving smooth gradients, we introduce geometrically decaying weights:
\begin{equation}
w_i
=
\frac{\max\!\left(\gamma^{\mathrm{pos}(i)},\, w_{\min}\right)}
{\sum_{j=1}^{N}
\max\!\left(\gamma^{\mathrm{pos}(j)},\, w_{\min}\right)},
\end{equation}
where $\gamma\in(0,1)$ controls decay and $w_{\min}>0$ prevents vanishing gradients in lower-ranked positions. Geometric decay provides a smooth approximation to Top-$K$ emphasis without introducing discontinuities. The resulting weighted KL objective is
\begin{equation}
\mathcal{L}_{\mathrm{list}}
=
\sum_{i=1}^{N}
w_i Q_i
\bigl(\log Q_i - \log P_i\bigr).
\end{equation}
This formulation prioritizes ranking precision in the head of the distribution while maintaining global consistency.

\paragraph{Auxiliary Regression Loss.}

To retain return magnitude information and reduce scale ambiguity under purely ranking-based optimization, we incorporate a pointwise regression term:
\begin{equation}
\mathcal{L}_{\mathrm{mse}}
=
\frac{1}{N}\sum_{i=1}^{N}(s_i - y_i)^2,
\end{equation}
where $y_i$ denotes realized next-period return. While $\mathcal{L}_{\mathrm{list}}$ governs cross-sectional ordering, $\mathcal{L}_{\mathrm{mse}}$ anchors score magnitudes to observed returns, improving numerical stability and supporting score-based allocation.

\paragraph{Graph Sparsity Regularization.}

Let $\mathbf{A}_\phi \in \mathbb{R}^{N \times N}$ denote the learned sparsified adjacency matrix,
where $(A_\phi)_{ij} = (A_0)_{ij} g_{ij}$ and $g_{ij} \in (0,1)$ are the learnable gating coefficients.
Although the NMI prior provides an information-theoretic initialization, 
unconstrained optimization may lead to overly dense connectivity or degenerate 
solutions in which most gates saturate toward 0 or 1. 
Such behavior can increase over-smoothing and reduce structural interpretability.

To encourage stable and well-conditioned graph structure, we introduce a sparsity-aware regularization term consisting of two complementary components
\begin{equation}
\mathcal{L}_{\mathrm{graph}}
=
\frac{1}{N(N-1)}
\sum_{i\ne j}(\mathbf{A}_\phi)_{ij}
+
\left(
\frac{1}{N(N-1)}
\sum_{i\ne j} g_{ij}
-
\rho
\right)^2,
\end{equation}
where the first term penalizes excessive average edge strength to control global connectivity magnitude and mitigates over-aggregation effects, and the second term constrains the expected gating density toward a target sparsity level $\rho \in (0,1)$.

The overall objective is
\begin{equation}
\mathcal{L}=\mathcal{L}_{\mathrm{list}}+\mathcal{L}_{\mathrm{mse}}+\mathcal{L}_{\mathrm{graph}}.
\end{equation}

\section{Experiments}
\subsection{Data preprocessing}

We construct a panel dataset of the 50 largest $S\&P$ 500 constituents by market capitalization at the start of the sample period to mitigate survivorship bias. The sample covers 752 NYSE trading days from 2022-01-03 to 2024-12-30. Daily OHLCV data are collected from Yahoo Finance and aligned with the NYSE trading calendar.

Features are winsorized at three standard deviations and standardized using training-set statistics to prevent information leakage. The data are split chronologically into training, validation, and test sets (8:1:1). Daily log returns from adjusted closing prices serve as prediction targets: returns are cross-sectionally standardized for ranking, while log returns are used for regression. Standardized features are arranged into rolling windows of length $L$, forming tensors of shape $[N, L, F]$. For each window ending at time $t$, the model predicts the return at $t+1$. Additional preprocessing details are provided in Appendix.

\subsection{Baseline Models}

To provide a structured comparison, we adopt baseline models covering classical statistical forecasting, deep sequential encoders, and graph-based spatio-temporal architectures. ARIMA is used as a representative linear time-series benchmark \citep{sawhney2021exploring, feng2022relation, zheng2023relational}. GRU and LSTM serve as recurrent neural network baselines for nonlinear temporal modeling and are widely used in financial prediction studies \citep{xu2021hist, he2022static}. 

 To incorporate cross-sectional dependencies, we further extend the temporal backbone with graph operators. GRU-GCN augments GRU with a graph convolutional layer using the same pre-computed NMI graph to ensure a fair comparison under a shared relational prior. GRU-GAT replaces the GCN layer with graph attention to enable adaptive neighbor weighting and maintain consistency with the GAT operator used in our proposed model. Together, these baselines provide a representative benchmark suite covering major paradigms in recent stock prediction and selection research. Please see the Appendix for implementation details.

\subsection{Evaluation}

To evaluate model performance, we consider both predictive accuracy and investment profitability. Ranking quality is evaluated using Mean Reciprocal Rank (MRR), which measures how highly the true top-performing stock appears in the predicted ranking \citep{gao2021graph, feng2022relation, zheng2023relational}. We also report Rank-Biased Overlap (RBO) to assess the similarity between predicted and ground-truth ranking lists, which is particularly relevant for portfolio construction \citep{saha2021stock}. In addition, Mean Squared Error (MSE) is included to measure prediction accuracy of return values \cite{gao2021graph, feng2022relation}. To evaluate investment performance, we conduct daily backtesting by constructing a long-only Top-5 portfolio from cross-sectional model scores. Portfolios are rebalanced daily (close-to-close), weighted via softmax allocation, and evaluated using net-of-fee returns with proportional turnover costs. Performance is reported using cumulative return (IRR), annualized Sharpe Ratio, and maximum drawdown (MDD), capturing profitability, risk-adjusted return, and downside risk. Please see the Appendix for implementation details.

\begin{table}[t]
\centering
\caption{\small Performance of STN-TGAT and baseline models. The best results are in \textbf{bold}.Higher is better for MRR, RBO, IRR, and Sharpe; lower is better for MSE and MDD.  check the left tables}\label{tab:baseline_perf}
\resizebox{\textwidth}{!}{%
\begin{tabular}{lcccccc}
\toprule
\textbf{MODEL} & ARIMA & GRU & LSTM & GRU-GCN & GRU-GAT & STN-TGAT \\
\midrule
MRR         & 0.1079 $\pm$ 0.0000 & 0.2120 $\pm$ 0.0059 & \textbf{0.2222 $\pm$ 0.0119} & 0.0768 $\pm$ 0.0332 & 0.1760 $\pm$ 0.0127 & 0.1879 $\pm$ 0.0345 \\
RBO         & 0.0316 $\pm$ 0.0000 & 0.0580 $\pm$ 0.0014 & \textbf{0.0601 $\pm$ 0.0024} & 0.0193 $\pm$ 0.0117 & 0.0468 $\pm$ 0.0037 & 0.0509 $\pm$ 0.0118 \\
MSE         & \textbf{0.0017 $\pm$ 0.0000} & 0.5142 $\pm$ 0.2321 & 0.2720 $\pm$ 0.4558 & 1.0727 $\pm$ 1.3167 & 0.5013 $\pm$ 0.3264 & 0.3115 $\pm$ 0.4176 \\
\midrule
IRR         & 0.0426 $\pm$ 0.0000 & 0.1341 $\pm$ 0.0359 & 0.1143 $\pm$ 0.0377 & -0.0435 $\pm$ 0.0166 & -0.0012 $\pm$ 0.0347 & \textbf{0.1807 $\pm$ 0.1176} \\
Sharpe      & 1.1145 $\pm$ 0.0000 & 2.6553 $\pm$ 0.5632 & 2.1899 $\pm$ 0.6187 & -1.4598 $\pm$ 0.6272 & 0.0035 $\pm$ 0.7847 & \textbf{2.9940 $\pm$ 1.3287} \\
MDD         & \textbf{0.0456 $\pm$ 0.0066} & 0.0585 $\pm$ 0.0099 & 0.0719 $\pm$ 0.0086 & 0.0799 $\pm$ 0.0075 & 0.0799 $\pm$ 0.0075 & 0.0634 $\pm$ 0.0130 \\
\bottomrule
\end{tabular}%
}
\end{table}

\section{Experiment results}
\subsection{Ranking Accuracy and Investment Performance Comparision}


Table~\ref{tab:baseline_perf} reports the comparative performance of all baseline models and the proposed STN-TGAT framework. In terms of ranking accuracy,STN-TGAT achieves competitive results, with an MRR of 0.1879 and an RBO of 0.0509. Although LSTM attains slightly higher values on certain pure ranking metrics, STN-TGAT maintains stable top-rank consistency while extracting signals that are more effective for downstream portfolio construction.

\begin{figure}[t]
  \centering
  \begin{subfigure}[t]{0.32\textwidth}
    \centering
    \includegraphics[width=\linewidth]{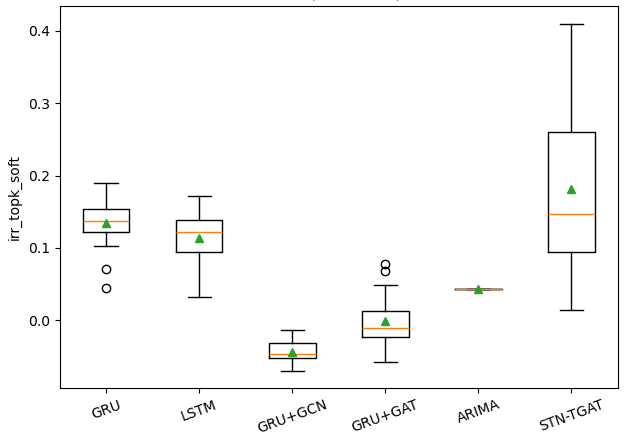}
    \caption{\small IRR}
    \label{fig:irr_box}
  \end{subfigure}
  \hfill
  \begin{subfigure}[t]{0.32\textwidth}
    \centering
    \includegraphics[width=\linewidth]{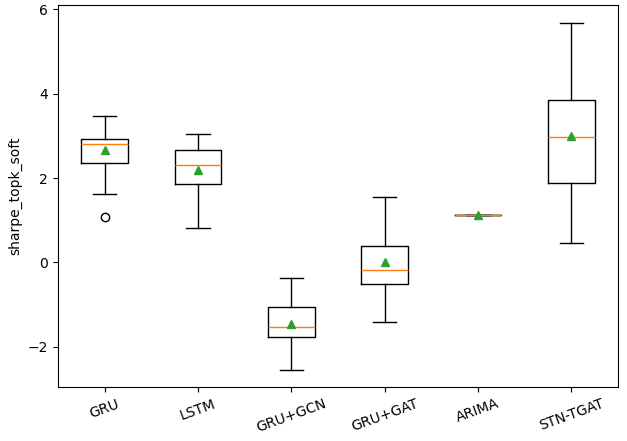}
    \caption{\small Sharpe ratio}
    \label{fig:sharpe_box}
  \end{subfigure}
  \hfill
  \begin{subfigure}[t]{0.32\textwidth}
    \centering
    \includegraphics[width=\linewidth]{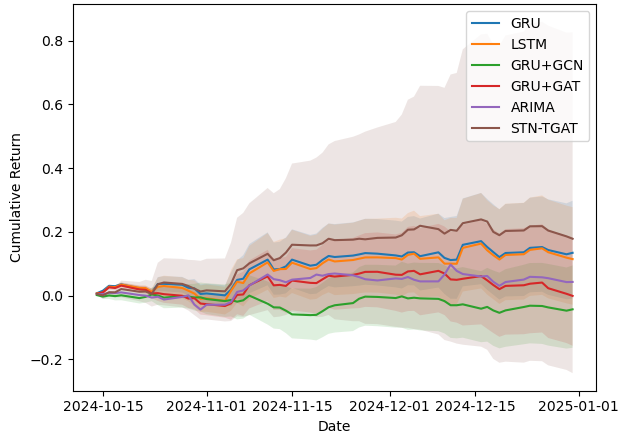}
    \caption{\small Cumulative return}
    \label{fig:cum_return}
  \end{subfigure}
  
  \caption{\small Performance comparison of different methods.}
  \label{fig:baseline_comparison}
\end{figure}

The superiority of STN-TGAT becomes more evident in the backtesting results. As reported in Table~\ref{tab:baseline_perf}, STN-TGAT achieves the highest IRR (18.07\%) and Sharpe ratio (2.99), outperforming the strongest recurrent baseline GRU. The distribution plots in Fig.~\ref{fig:irr_box} and Fig.~\ref{fig:sharpe_box} further confirm that STN-TGAT consistently yields higher median returns and stronger risk-adjusted performance compared with other models. Moreover, Fig.~\ref{fig:cum_return} illustrates the cumulative return trajectories during the test period. STN-TGAT maintains the highest equity curve throughout the evaluation window, demonstrating its ability to sustain portfolio growth even during volatile market phases. Despite operating in noisy financial environments, STN-TGAT exhibits robust risk characteristics. As shown in Table~\ref{tab:baseline_perf}, it achieves lower maximum drawdown than most neural baselines, indicating improved downside protection.

Overall, the evaluation considers both ranking accuracy and after-fee portfolio performance. While no model dominates all metrics, STN-TGAT achieves the most consistent balance, translating reliable Top-5 selection into stronger backtested returns with a competitive risk profile under transaction costs. These results highlight its practical suitability for real-world portfolio construction.

\subsection{Ablation Studies}
\subsubsection{Loss function}

\begin{table}[t]
\centering
\caption{\small Ablation study of loss.} \label{tab:ablation_cols}
\resizebox{\textwidth}{!}{%
\begin{tabular}{lccccc}
\toprule
Metric &
ListNet+MSE &
ListNet-only &
MSE-only &
Spearman+MSE &
Pairwise+MSE \\
\midrule

RBO     & 0.0516 $\pm$ 0.0127 & 0.0519 $\pm$ 0.0135 & 0.0372 $\pm$ 0.0077 & 0.0487 $\pm$ 0.0048 & 0.0407 $\pm$ 0.0076 \\
MRR     & 0.1875 $\pm$ 0.0378 & 0.1904 $\pm$ 0.0401 & 0.1263 $\pm$ 0.0195 & 0.1643 $\pm$ 0.0245 & 0.1583 $\pm$ 0.0168 \\
\midrule
IRR     & 0.2132 $\pm$ 0.1309 & 0.2161 $\pm$ 0.1326 & 0.0395 $\pm$ 0.0730 & 0.0945 $\pm$ 0.0145 & 0.0646 $\pm$ 0.0543 \\
Sharpe  & 3.3130 $\pm$ 1.5240 & 3.3160 $\pm$ 1.5250 & 0.7570 $\pm$ 1.7100 & 2.3560 $\pm$ 0.3300 & 1.4740 $\pm$ 1.0780 \\
MDD     & 0.0618 $\pm$ 0.0136 & 0.0646 $\pm$ 0.0142 & 0.0779 $\pm$ 0.0122 & 0.0744 $\pm$ 0.0058 & 0.0812 $\pm$ 0.0227 \\

\bottomrule
\end{tabular}%
}
\end{table}

We conduct an ablation study to assess the contribution of each component in the proposed training objective for Top-5 portfolio construction. Our final loss for prediction combines a global top-weighted ListNet term with an auxiliary MSE term. To isolate their contributions, we evaluate \emph{ListNet-only} and \emph{MSE-only} variants. In addition, two alternative ranking objectives, \emph{Spearman} \citet{hu2025study} and \emph{Pairwise} \cite{patel2024systematic}, are tested in combination with MSE.

As shown in Table~\ref{tab:ablation_cols}, models without the ListNet component (\emph{MSE-only}, \emph{Spearman+MSE}, \emph{Pairwise+MSE}) show substantially weaker investment performance. \emph{MSE-only} model achieves an IRR of 0.0395 and Sharpe ratio of 0.7570, far below the 0.2132 IRR and 3.3130 Sharpe obtained by \emph{ListNet+MSE}. Similarly, \emph{Spearman+MSE} and \emph{Pairwise+MSE} yield lower IRRs (0.0945 and 0.0646), indicating that listwise ranking supervision is essential for decision-aligned stock selection. While \emph{ListNet-only} already achieves strong profitability, incorporating the auxiliary MSE term (\emph{ListNet+MSE}) leads to the most robust portfolio performance, achieving comparable ranking metrics with improved risk-adjusted returns and lower drawdowns (0.0618 vs.\ 0.0646). This suggests that the MSE term acts as a calibration and regularization component, stabilizing score magnitudes and improving the reliability of portfolio outcomes.

Overall, the results indicate that listwise supervision is critical for investable performance, while the hybrid objective provides a balanced and practically viable training strategy.

\subsubsection{Effectiveness of the Prior Knowledge}

\begin{table}[t]
\centering
\caption{\small Ablation study results across different model configurations. }
\label{tab:ablation_models}
\resizebox{\textwidth}{!}{%
\begin{tabular}{lccccc}
\toprule
\textbf{MODEL} &
Transformer-GAT (no prior) &
Transformer-GAT (no sparsity) &
Transformer-GAT (GLASSO) &
STN-TGAT \\
\midrule
RBO        & 0.0413 $\pm$ 0.0053 & 0.0413 $\pm$ 0.0056 & 0.0406 $\pm$ 0.0072 & \textbf{0.0509 $\pm$ 0.0118} \\
MRR        & 0.1521 $\pm$ 0.0214 & 0.1514 $\pm$ 0.0266 & 0.1462 $\pm$ 0.0226 & \textbf{0.1879 $\pm$ 0.0345} \\
MSE        & 0.5192 $\pm$ 0.3942 & \textbf{0.3073 $\pm$ 0.3239} & 0.6019 $\pm$ 0.5262 & 0.3115 $\pm$ 0.4176 \\
\midrule
IRR      & 0.0863 $\pm$ 0.0692 & 0.1152 $\pm$ 0.0551 & 0.0381 $\pm$ 0.0672 & \textbf{0.1807 $\pm$ 0.1176} \\
Sharpe   & 1.8932 $\pm$ 1.2710 & 2.5884 $\pm$ 0.9710 & 0.7480 $\pm$ 1.5070 & \textbf{2.9940 $\pm$ 1.3287} \\
MDD      & 0.0574 $\pm$ 0.0093 & \textbf{0.0520 $\pm$ 0.0116} & 0.0622 $\pm$ 0.0133 & 0.0634 $\pm$ 0.0130 \\
\bottomrule
\end{tabular}%
}

\end{table}

\begin{figure}[t]
  \centering
  \begin{subfigure}[t]{0.32\textwidth}
    \centering
    \includegraphics[width=\linewidth]{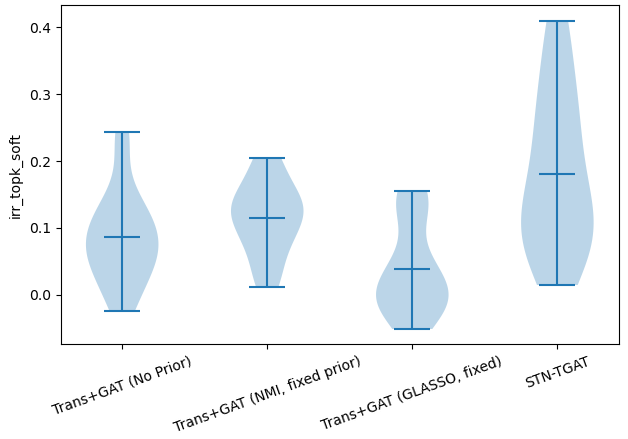}
    \caption{IRR }
    \label{fig:abl_irr}
  \end{subfigure}
  \hfill
  \begin{subfigure}[t]{0.32\textwidth}
    \centering
    \includegraphics[width=\linewidth]{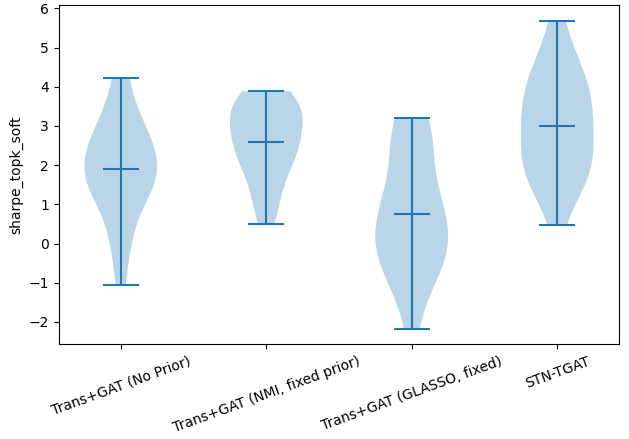}
    \caption{Sharpe}
    \label{fig:abl_sharpe}
  \end{subfigure}
  \hfill
  \begin{subfigure}[t]{0.32\textwidth}
    \centering
    \includegraphics[width=\linewidth]{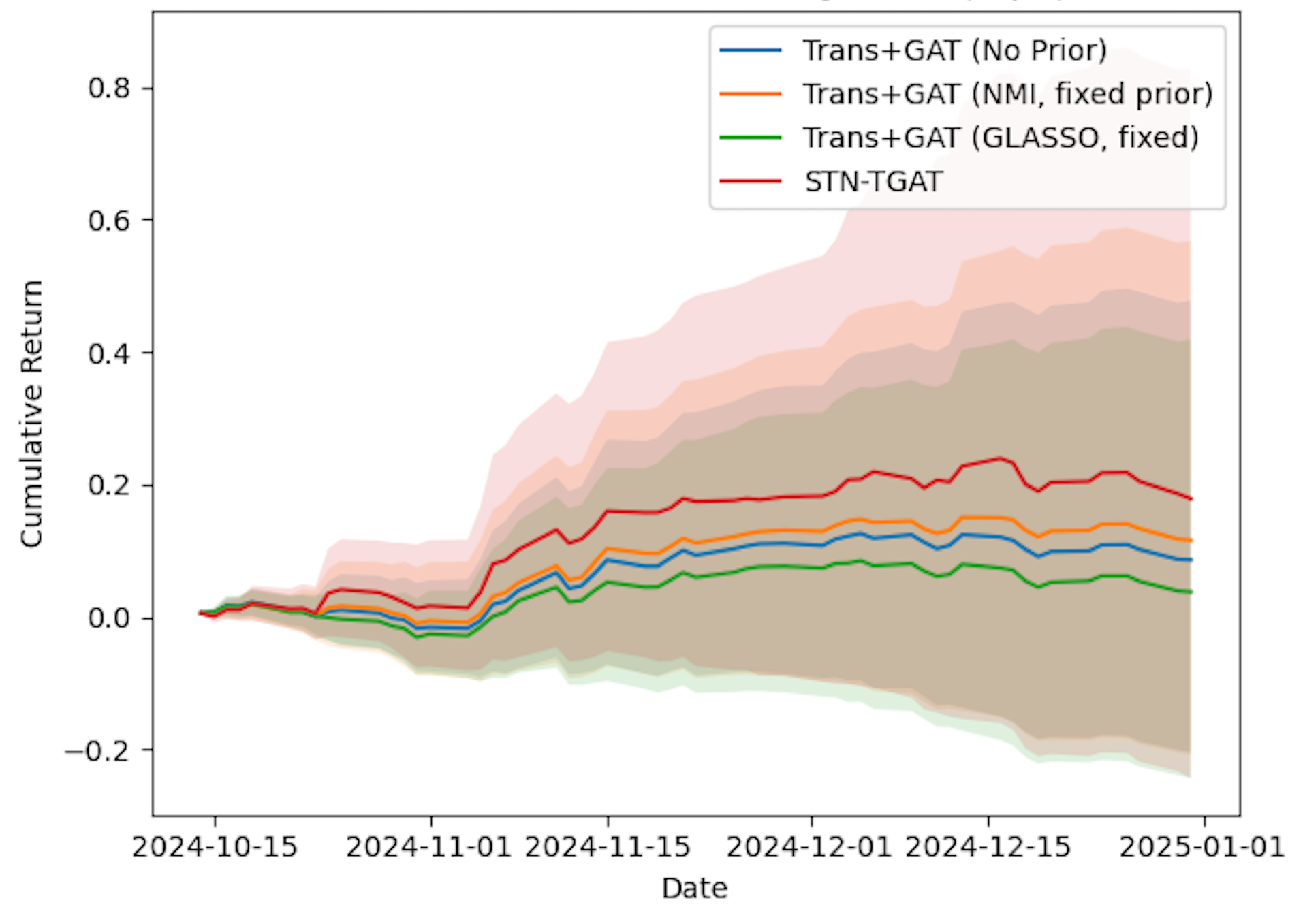}
    \caption{Cumulative Return}
    \label{fig:abl_cum}
  \end{subfigure}
  \caption{Ablation study visualization: (a) and (b) show the distributions of IRR and Sharpe ratios via violin plots; (c) displays the cumulative mean returns with shaded standard deviation bands.}
  \label{fig:ablation_visuals}
\end{figure}

To assess the contribution of each structural component, we ablate the relational prior, prior type, and adaptive sparsification. Removing the structural prior substantially reduces investment performance. The Transformer-GAT (no prior) variant shows a large drop in IRR from 18.07\% to 8.63\% and a decrease in Sharpe ratio from 2.9940 to 1.8932. (Table~\ref{tab:ablation_models}), with a visibly more dispersed return distribution (Fig.~\ref{fig:abl_irr}). Replacing the nonlinear NMI prior with a linear GLASSO prior further degrades results. The Transformer-GAT (GLASSO) produces the lowest IRR (0.0381) and Sharpe ratio (0.7480), and an almost flat cumulative return trajectory (Fig.~\ref{fig:abl_cum}), suggesting that the nonlinear prior captures dependencies more relevant for Top-$K$ selection. Finally, adaptive sparsification also plays an important role. Although the Transformer-GAT (no sparsity) model still benefits from the NMI prior, disabling sparsification introduces noisy edges, reducing the Sharpe ratio from 2.9940 to 2.5884. As shown in Fig.~\ref{fig:abl_sharpe}, STN-TGAT achieves a more concentrated high-return distribution, suggesting that learnable sparsification effectively filters redundant connections.

\begin{figure}[t]
    \centering
    \begin{subfigure}[b]{0.32\textwidth}
        \centering
        \includegraphics[width=\textwidth]{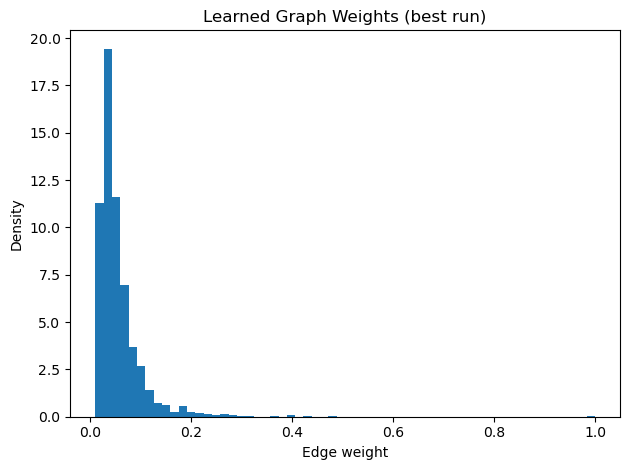}
        \caption{\small Learned edge-weight histogram.}
        \label{fig:5.6a}
    \end{subfigure}
    \hfill
    \begin{subfigure}[b]{0.32\textwidth}
        \centering
        \includegraphics[width=\textwidth]{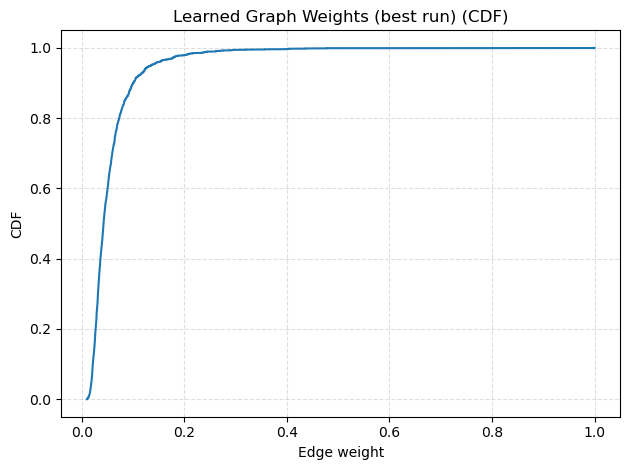}
        \caption{\small CDF of learned edge weights.}
        \label{fig:5.6b}
    \end{subfigure}
    \hfill
    \begin{subfigure}[b]{0.32\textwidth}
        \centering
        \includegraphics[width=\textwidth]{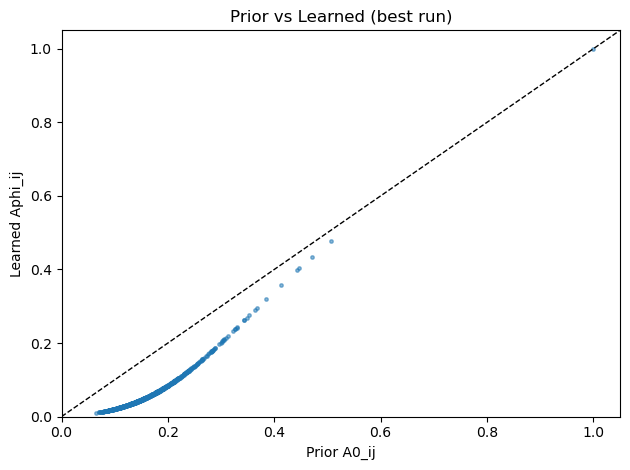}
        \caption{\small Prior vs.\ learned edge weights.}
        \label{fig:5.6c}
    \end{subfigure}
    \caption{\small Analysis of the learned adjacency. Panel (a) reports the distribution of learned edge weights, (b) shows the cumulative distribution, and (c) compares prior weights $A_0$ with learned weights $A_{\phi}$; the $y{=}x$ diagonal indicates perfect agreement.}
    \label{fig:5.6}
\end{figure}

\begin{figure}[t]
    \centering
    \begin{subfigure}[b]{0.48\textwidth}
        \centering
        \includegraphics[width=\textwidth]{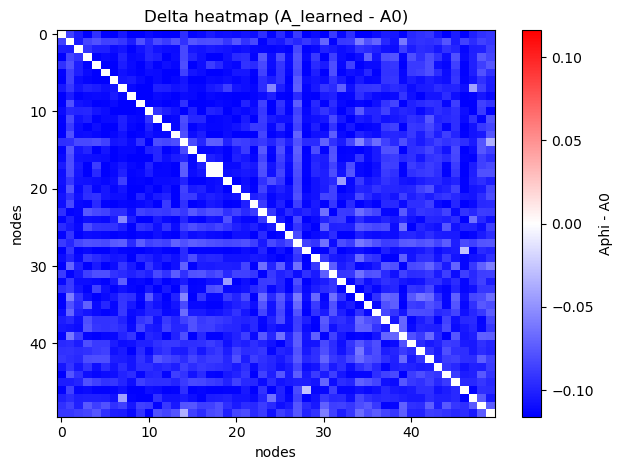}
        \caption{Delta heatmap: $A_{\phi}-A_{0}$}
        \label{fig:5.7a}
    \end{subfigure}
    \hfill
    \begin{subfigure}[b]{0.48\textwidth}
        \centering
        \includegraphics[width=\textwidth]{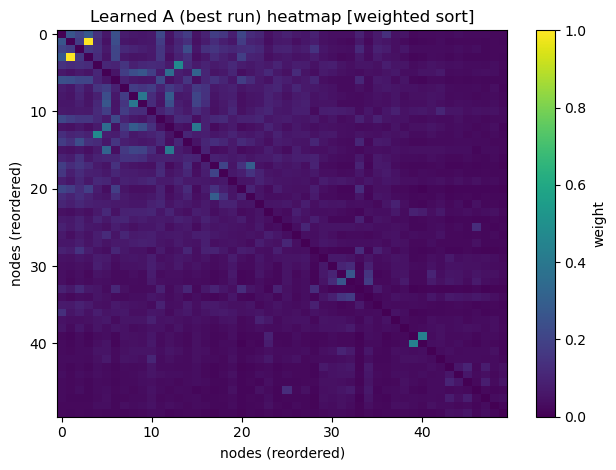}
        \caption{Learned $A_{\phi}$ heatmap (reordered)}
        \label{fig:5.7b}
    \end{subfigure}
    \caption{\small Heatmap diagnostics of the learned graph vs.\ the NMI prior. Panel (a) visualizes adjustments to the prior via the difference $A_{\phi}-A_{0}$, and (b) shows the learned adjacency after node reordering by aggregate edge weight.}
    \label{fig:5.7}
\end{figure}

The structural diagnostics are consistent with this observation. The learned edge weights exhibit a highly right-skewed distribution (Fig.~\ref{fig:5.6a}) and a steep CDF (Fig.~\ref{fig:5.6b}), indicating that most edges are assigned small magnitudes. The prior–learned comparison (Fig.~\ref{fig:5.6c}) and delta heatmap (Fig.~\ref{fig:5.7a}) show systematic down-weighting relative to the NMI prior, while the reordered adjacency (Fig.~\ref{fig:5.7b}) reveals concentrated high-weight regions alongside extensive low-weight areas. Together, these results suggest that the gating mechanism refines the prior topology by attenuating weaker connections, which aligns with the observed improvements in out-of-sample portfolio performance.

\subsubsection{Effectiveness of Top-$K$ Within-Set Weighting}

\begin{table}[t]
\centering
\caption{\small Effect of within-Top-5 allocation for STN-TGAT.}\label{tab:topk_weight_ablation}
\begin{tabular}{lcccc}
\toprule
Metric & & Top$K$-EqualWeight & & Top$K$-ScoreWeight \\
\midrule
IRR    & & 0.1521 $\pm$ 0.1119 & & 0.1807 $\pm$ 0.1176 \\
Sharpe & & 2.5595 $\pm$ 1.3979 & & 2.9940 $\pm$ 1.3287 \\
MDD    & & 0.0636 $\pm$ 0.0132 & & 0.0634 $\pm$ 0.0130 \\
\bottomrule
\end{tabular}
\end{table}

We further conduct an ablation study on the  within-Top-5 allocation strategy used in portfolio construction to assess whether translating model scores into conviction-aware weights provides additional practical benefits beyond pure stock selection. Specifically, keeping the trained STN-TGAT model and the Top-5 selection rule unchanged, we compare two allocation strategies within the selected Top-5 set: (i) \emph{equal-weighting}, which assigns identical weights to all selected stocks, and (ii) \emph{score-based weighting}, which allocates weights according to the relative magnitudes of predicted scores from Eq. \ref{eq:score_head}, thereby reflecting the model's confidence when forming daily portfolios.

As shown in Table~\ref{tab:topk_weight_ablation}, the score-based weighting scheme consistently improves portfolio-level performance compared with equal-weighting. Specifically, the weighted strategy achieves higher IRR (0.1807 vs.\ 0.1521) and Sharpe ratio (2.9940 vs.\ 2.5595), while maintaining comparable maximum drawdown (0.0634 vs.\ 0.0636). Although the standard deviations indicate variability across runs, the consistent improvement in mean return and risk-adjusted performance suggests that the predicted score magnitudes contain information beyond ranking order alone. Overall, these findings indicate that incorporating score-based allocation within the selected Top-5 set can enhance out-of-sample portfolio performance under the same selection rule, supporting the practical utility of the model’s output for both ranking and capital allocation.

\section{Conclusion}

This paper presents the Soft-Thresholded NMI-prior Transformer Graph Attention Network (STN-TGAT), a spatial-temporal framework for Top-$K$ equity selection. We formulate daily stock selection as a decision-aligned ranking problem and introduce a top-weighted ListNet objective that emphasizes accuracy within the investable set. An auxiliary MSE term anchors predicted score magnitudes to realized returns, linking predictive performance to economically meaningful portfolio results. The architecture integrates Transformer-based temporal attention with graph attention mechanisms to jointly capture long-range temporal dependencies and dynamic cross-sectional interactions. In addition, a soft-thresholded Normalized Mutual Information (NMI) prior guides graph construction through adaptive sparsification, systematically attenuating weaker connections while preserving stronger relational structure.

The framework is evaluated on real-world $S\&P$ 500 data under a Top-5 selection setting with explicit transaction costs. Across multiple ablation configurations, STN-TGAT achieves competitive ranking performance and improved risk-adjusted returns relative to alternative structural variants. These results indicate that integrating nonlinear relational priors with adaptive gating to a coherent and practically effective approach to portfolio construction.

\bibliographystyle{plainnat}
\bibliography{references}

\end{document}